\pdfoutput=1
\documentclass[11pt,a4paper]{article}
\usepackage[hyperref]{naaclhlt2019}
\usepackage{times}
\usepackage{latexsym}
\usepackage{color}
\usepackage{amsmath}
\usepackage{amssymb}
\usepackage{multirow}
\usepackage{pifont}
\usepackage{longtable}

\newcommand{\qt}[1]{{\textquoteleft {#1}\textquoteright}}

\renewcommand{\vec}[1]{\mathbf{#1}}
\DeclareMathOperator*{\argmax}{arg\,max}

\definecolor{UniBlue}{RGB}{0, 132, 180}
\definecolor{clearBlue}{RGB}{230,242,243}

\aclfinalcopy 

\title{Harry Potter and the Action Prediction Challenge from Natural Language}

\author{
  David Vilares\\
  Universidade da Coru\~{n}a, CITIC \\
  Departamento de Computaci\'{o}n \\
  Campus de Elvi\~{n}a s/n, 15071 \\ A Coru\~{n}a, Spain \\
  {\tt david.vilares@udc.es} \\
  \\\And
  Carlos G\'{o}mez-Rodr\'{i}guez \\
  Universidade da Coru\~{n}a, CITIC \\
  Departamento de Computaci\'{o}n \\
  Campus de Elvi\~{n}a s/n, 15071 \\ A Coru\~{n}a, Spain \\
  {\tt carlos.gomez@udc.es}  \\}

\date{}

\begin{document}
\maketitle
\begin{abstract}

We explore the challenge of action prediction from textual descriptions of scenes, a testbed to approximate whether text inference can be used to predict upcoming actions.
As a case of study, we consider the world of the Harry Potter fantasy novels and inferring what spell will be cast next given a fragment of a story. 
Spells act as keywords that abstract actions (e.g. \qt{Alohomora} to open a door) and denote a response to the environment. This idea is used to automatically build \textsc{hpac}, a corpus containing 82\,836 samples and 85 actions. 
We then evaluate different baselines.
Among the tested models, an \textsc{lstm}-based approach obtains the best performance for frequent actions and large scene descriptions, but approaches such as logistic regression behave well on infrequent actions.

\end{abstract}

\section{Introduction}

Natural language processing (\textsc{nlp}) has achieved significant advances in reading comprehension tasks~\cite{chen-bolton-manning:2016:P16-1,2017arXiv171203609S}. These are partially due to embedding methods \cite{mikolov2013distributed,devlin2018bert} and neural networks~\cite{rosenblatt1958perceptron,hochreiter1997long,vaswani2017attention}, but also to the availability of new resources and challenges. For instance, in cloze-form tasks~\cite{hermann2015teaching,bajgar2016embracing}, the goal is to predict the missing word given a short context. \newcite{weston2015towards} presented baBI, a set of proxy tasks for reading comprenhension. In the SQuAD corpus \cite{squad}, the aim is to answer questions given a Wikipedia passage. \newcite{2017arXiv171207040K} introduce NarrativeQA, where answering the questions requires to process entire stories. In a related line, \newcite{2017arXiv171011601F} use fictional crime scene investigation data, from the CSI series, to define a task where the models try to answer the question: \qt{who committed the crime?}.

In an alternative line of work, script induction \cite{schank1977scripts} has been also a useful approach to evaluate inference and semantic capabilities of \textsc{nlp} systems. Here, a model processes a document to infer new sequences that reflect events that are statistically probable (e.g. go to a restaurant, be seated, check the menu, \dots).  For example, \newcite{chambers2008unsupervised} introduce narrative event chains, a representation of structured knowledge of a set of events occurring around a protagonist. They then propose a method to learn statistical scripts, and also introduce two different evaluation strategies. With a related aim, \newcite{Pichotta2014Statistical} propose a multi-event representation of statistical scripts to be able to consider multiple entities. These same authors \cite{Pichotta2016Using} have also studied the abilities of recurrent neural networks for learning scripts, generating upcoming events given a raw sequence of tokens, using \textsc{bleu} \cite{papineni2002bleu} for evaluation.

This paper explores instead a new task: action prediction from natural language descriptions of scenes.
The challenge is addressed as follows: given a natural language input sequence describing the scene, such as a piece of a story coming from a transcript, the goal is to infer which action is most likely to happen next. 

\paragraph{Contribution} 
We introduce a fictional-domain English corpus set in the world of Harry Potter novels. 
The domain is motivated by the existence of a variety of spells in these literary books, associated with keywords that can be seen as unambiguous markers for actions that potentially relate to the previous context. This is used to automatically create a natural language corpus coming from hundreds of users, with different styles, interests and writing skills. 
We then train a number of standard baselines to predict upcoming actions, a task that requires to be aware of the context. 
In particular, we test a number of generic models, from a simple logistic regression to neural models. Experiments shed some light about their strengths and weaknesses and how these are related to the frequency of each action, the existence of other semantically related actions and the length of the input story.

\section{HPAC: The Harry Potter's Action prediction Corpus}\label{section-HPAC}

To build an action prediction corpus, we need to: (1) consider the set of actions, and (2) collect data where these occur. Data should come from different users, to approximate a real natural language task.
Also, it needs to be annotated, determining that a piece of text  ends up triggering an action.
These tasks are however time consuming, as they require annotators to read vast amounts of large texts.
In this context, machine comprehension resources usually establish a compromise between their complexity and the costs of building them \cite{hermann2015teaching,2017arXiv171207040K}.

\begin{table*}[h]
\small
\begin{center}
\tabcolsep=0.08cm
\begin{tabular}{|p{14cm}|p{1.5cm}|}
\hline
 \bf Text fragment & \bf Action  \\ 
\hline

Ducking under Peeves, they ran for their lives, right to the end of the corridor where they slammed into a  door - and it was locked. \qt{This is it!} Ron moaned, as they pushed helplessly at the door, \qt{We're done for! This is the end!} They could hear footsteps, Filch running as fast as he could toward Peeves's shouts. \qt{Oh, move over}, Hermione snarled. She grabbed Harry's wand, tapped the lock, and whispered, \qt{\bf Alohomora}. & Unlock the door\\

\hline

And then, without warning, Harry's scar exploded with pain. It was agony such as he had never felt in all his life; his wand slipped from his fingers as he put  his hands over his face; his knees buckled; he was on the ground and he could see nothing at all; his head was about to split open. From far away, above his head, he heard a high, cold voice say, \qt{Kill the spare.} A swishing noise and a second voice, which screeched  the words to the night: \qt{\bf Avada Kedavra} & Kill a target \\ 

\hline

Harry felt himself being pushed hither and thither by people whose faces he could not see. Then he heard Ron yell with pain. \qt{What happened?} said Hermione anxiously, stopping so abruptly that Harry walked into her. \qt{Ron, where are you? Oh, this is stupid} - \qt{\bf Lumos} & Turn on a light\\

\hline

\end{tabular}
\end{center}
\caption{\label{spells-examples} Examples from the Harry Potter books showing how spells map to reactions to the environment.}
\end{table*}

\subsection{Domain motivation}

We rely on an intuitive idea that uses transcripts from the Harry Potter world to build up a corpus for textual action prediction. The domain has a set of desirable properties to evaluate reading comprehension systems, which we now review.

Harry Potter novels define a variety of spells. These are keywords cast by witches and wizards to achieve purposes, such as turning on a light (\qt{Lumos}), unlocking a door (\qt{Alohomora}) or killing (\qt{Avada Kedavra}). They abstract complex and non-ambiguous actions. Their use also makes it possible to build an automatic and self-annotated corpus for action prediction. The moment a spell occurs in a text
represents a response to the environment, and hence, it can be used to label the preceding text fragment as a scene description that ends up triggering that action. Table \ref{spells-examples} illustrates it with some examples from the original books.

This makes it possible to consider texts from the magic world of Harry Potter as the domain for the action prediction corpus, and the spells as the set of eligible actions.\footnote{Note that the corpus is built in an automatic way and some occurrences might not correspond to actions, but for example, to a description of the spell or even some false positive samples. Related to this, we have not censored the content of the stories, so some of them might contain adult content.} Determining the length of the preceding context, namely \emph{snippet}, that has to be considered as the scene description is however not trivial. 
This paper considers experiments (\S \ref{section-results}) using snippets with the 32, 64, 96 and 128 previous tokens to an action. We provide the needed scripts to rebuild the corpus using arbitrary lengths.\footnote{\url{https://github.com/aghie/hpac}}

\subsection{Data crawling}

The number of occurrences of spells in the original Harry Potter books is small (432 occurrences), which makes it difficult to train and test a machine learning model. However, the amount of available fan fiction for this saga allows to create a large corpus. For \textsc{hpac}, we used fan fiction (and \emph{only} fan fiction texts) from \url{https://www.fanfiction.net/book/Harry-Potter/} and a version of the crawler by \newcite{milli2016beyond}.\footnote{Due to the website's Terms of Service, the corpus cannot be directly released.
} 
We collected Harry Potter stories written in English and marked with the status \qt{completed}. From these we extracted a total of 82\,836 spell occurrences, that we used to obtain the scene descriptions. Table \ref{corpus-statistics} details the statistics of the corpus (see also Appendix \ref{appendix-corpus-details}). Note that similar to Twitter corpora, fan fiction stories can be deleted over time by users or admins, causing losses in the dataset.\footnote{They also can be modified, making it unfeasible to retrieve some of the samples.}

\paragraph{Preprocessing} 
We tokenized the samples with \cite{manning2014stanford} and merged the occurrences of multi-word spells into a single token.

\begin{table}[bpth]
\small
\begin{center}
\tabcolsep=0.08cm
\begin{tabular}{|l|r|r|r|}
\hline
 \bf Statistics& \bf Training & \bf Dev & \bf Test \\
 \hline
 \#Actions  &  85 & 83& 84 \\ 
 \#Samples &  66\,274 & 8\,279& 8\,283\\
 \#Tokens (s=32) & 2\,111\,180&263\,573&263\,937\\
 \#Unique tokens (s=32) &33\,067&13\,075&13\,207\\
 \#Tokens (s=128) & 8\,329\,531
 &1\,040\,705&1\,041\,027\\
 \#Unique tokens (s=128) &60\,379&25\,146&25\,285\\
\hline

\end{tabular}
\end{center}
\caption{\label{corpus-statistics} Corpus statistics: $s$  is the length of the snippet.}
\end{table}

\section{Models}\label{section-methods}

This work addresses the task as a classification problem, and in particular as a sequence to label classification problem. For this reason, we rely on standard models used for this type of task: multinomial logistic regression, a multi-layered perceptron, convolutional neural networks and long short-term memory networks. 
We outline the essentials of each of these models, but will treat them as black boxes. In a related line, \newcite{kaushik2018much} discuss the need of providing rigorous baselines that help better understand the improvement coming from future and  complex models, and also the need of not demanding architectural novelty when introducing new datasets. 

Although not done in this work, an alternative (but also natural) way to address the task is as a special case of language modelling, where the output vocabulary is restricted to the size of the `action' vocabulary. Also, note that the performance for this task is not expected to achieve a perfect accuracy, as there may be situations where more than one action is reasonable, and also because writers tell a story playing with elements such as surprise or uncertainty.

The source code for the models can be found in the GitHub repository mentioned above.

\paragraph{Notation} $w_{1:n}$ denotes a sequence of words $w_1,...,w_n$ that represents the scene, with $w_i \in V$. $F_\theta(\cdot)$ is a function parametrized by $\theta$. The task is cast as $F: V^n \rightarrow A$, where $A$ is the set of actions.

\subsection{Machine learning models}\label{section-ml-models}

The input sentence $w_{1:n}$ is encoded as a one-hot vector, $\vec{v}$ (total occurrence weighting scheme).

  \paragraph{Multinomial Logistic Regression} 
 Let \textsc{mlr}$_\theta(\vec{v})$ be an abstraction of a multinomial logistic regression parametrized by $\theta$, the output for an input $\vec{v}$ is computed as the $\argmax_{a \in A}$ $P(y=a|\vec{v})$, where $P(y=a|\vec{v})$ is a $softmax$ function, i.e, $P(y=a|\vec{v}) = \frac{e^{W_{a} \cdot \vec{v}}} {\sum_{a'}^{A} e^{W_{a'} \cdot \vec{v}}}$.

\paragraph{MultiLayer Perceptron} We use one hidden layer with a rectifier activation function ($relu(x)$=$max(0,x)$). 
The output is computed as \textsc{mlp}$_\theta(\vec{v})$= $softmax(W_2 \cdot relu(W \cdot \vec{v} + \vec{b}) + \vec{b_2})$.

\subsection{Sequential models}

The input sequence is represented as a sequence of word embeddings, $\vec{w}_{1:n}$, where $\vec{w}_i$ is a concatenation of an internal embedding learned during the training process for the word $w_i$, and a pre-trained embedding extracted from GloVe \cite{pennington2014glove}\footnote{\url{http://nlp.stanford.edu/data/glove.6B.zip}}, that is further fine-tuned.

\paragraph{Long short-term memory network} \cite{hochreiter1997long}:
The output for an element $\vec{w}_i$ also depends on the output of $\vec{w}_{i-1}$. 
The \textsc{lstm}$_\theta(\vec{w}_{1:n})$\footnote{$n$ is set to be equal to the length of the snippet.} takes as input a sequence of word embeddings and produces a sequence of hidden outputs, $\vec{h}_{1:n}$ ($\vec{h}_{i}$ size set to 128). The last output of the \textsc{lstm}$_\theta$, $\vec{h}_n$, is fed to a \textsc{mlp}$_\theta$.

\paragraph{Convolutional Neural Network} \cite{lecun1995convolutional,kim2014convolutional}. 
It captures local properties over continuous slices of text by applying a convolution layer made of different filters. We use a wide convolution, with a window slice size of length 3 and 250 different filters.
The convolutional layer uses a $\mathit{relu}$ as the activation function. The output is fed to a max pooling layer, whose output vector is passed again as input to a \textsc{mlp}$_\theta$.

\section{Experiments}\label{section-results}

\paragraph{Setup} All \textsc{mlp}$_\theta$'s have 128 input neurons and 1 hidden layer. We trained up to 15 epochs using mini-batches (size=16), Adam (lr=$0.001$) \cite{kingma2014adam} and early stopping.\\

Table \ref{table-f-scores} shows the macro and weighted F-scores for the models considering different snippet sizes.\footnote{As we have addressed the task as a classification problem, we will use precision, recall and F-score as the evaluation metrics.} To diminish the impact of random seeds and local minima in neural networks, results are averaged across 5 runs.\footnote{Some macro F-scores do not lie within the Precision and Recall due to this issue.} `Base' is a majority-class model that maps everything to \qt{Avada Kedavra}, the most common action in the training set. This helps test whether the models predict above chance performance. When using short snippets (size=32), disparate models such as our \textsc{mlr}, \textsc{mlp} and \textsc{lstm}s achieve a similar performance. As the snippet size is increased, the \textsc{lstm}-based approach shows a clear improvement on the weighted scores\footnote{For each label, we compute their average, weighted by the number of true instances for each label. The F-score might be not between precision and recall.}, something that happens only marginally for the rest. However, from Table \ref{table-f-scores} it is hard to find out what the approaches are actually learning to predict.

\begin{table}[hbtp]
\small
\begin{center}
\tabcolsep=0.14cm
\begin{tabular}{|c|lrrrrrr|}
\hline
 \multirow{2}{*}{\bf Snippet} & \multirow{2}{*}{\bf Model} & \multicolumn{3}{c}{\bf Macro} & \multicolumn{3}{c|}{\bf Weighted} \\
 & & \bf P & \bf R & \bf F & \bf P & \bf R & \bf F \\
 \hline
 -&{Base} & 0.1 & 1.2 & 0.2 & 1.3 & 11.5 & 2.4\\ 
 \hline
 \multirow{4}{*}{32} & \textsc{mlr} &18.7&11.6&13.1&28.9&31.4&28.3\\
 & \textsc{mlp}  &19.1&9.8&10.3&\bf 31.7&32.1&28.0\\
 & \textsc{lstm}  &13.7&9.7&9.5&29.1&32.2&28.6\\
 & \textsc{cnn} &9.9&7.8&7.3&24.6&29.2&24.7\\
 \hline
 \multirow{4}{*}{64} & \textsc{mlr}  &\bf 20.6&12.3&13.9&29.9&32.1&29.0\\
 & \textsc{mlp}  &17.9&9.5&9.8&31.2&32.7&27.9\\
  & \textsc{lstm} &13.3&10.3&10.2&30.3&33.9&30.4\\

  & \textsc{cnn} &9.8&7.8&7.4&25.0&29.9&25.4\\
 \hline
  \multirow{4}{*}{96} & \textsc{mlr}  &20.4&\bf 13.3&\bf 14.6&30.3&32.0&29.3\\
 & \textsc{mlp}  &16.9&9.5&9.8&30.2&32.6&27.8\\
  & \textsc{lstm} &14.0&10.5&10.3&30.6&34.5&30.7\\
  & \textsc{cnn} &10.2&7.1&6.9&25.2&29.4&24.4\\
 \hline
  \multirow{4}{*}{128} & \textsc{mlr}  &19.6&12.1&12.9&30.0&31.7&28.2\\
 & \textsc{mlp}  &18.9&9.9&10.3& 31.4&32.9&28.0\\
 & \textsc{lstm}  &14.4&10.5&10.5&31.3&\bf35.1&\bf 31.1\\
 & \textsc{cnn}  &8.8&7.8&7.1&24.8&30.2&25.0\\

\hline

\end{tabular}
\end{center}
\caption{\label{table-f-scores} Macro and weighted F-scores over 5 runs.}
\end{table}

To shed some light, Table \ref{table-recall-at-k} shows their  performance according to a ranking metric, recall at $k$. The results show that the \textsc{lstm}-based approach is the top performing model, but the \textsc{mlp} obtains just slightly worse results. Recall at 1 is in both cases low, which suggests that the task is indeed complex and that using just \textsc{lstm}s is not enough. It is also possible to observe that even if the models have difficulties to correctly predict the action as a first option, they develop certain sense of the scene and consider the right one among their top choices. Table \ref{table-freq-infreq} delves into this by splitting the performance of the model into infrequent and frequent actions (above the average, i.e. those that occur more than 98 times in the training set, a total of 20 actions).
There is a clear gap between the performance on these two groups of actions, with a $\sim$50 points difference in recall at $5$. Also, a simple logistic regression performs similar to the \textsc{lstm} on the infrequent actions.

\begin{table}[hbtp]
\small
\begin{center}
\begin{tabular}{|c|ccccc|}
\hline
 \bf Snippet & \bf Model & \bf R@1  & \bf R@2 & \bf R@5 & \bf R@10  \\
 \hline
 -&{Base} & 11.5 & - & - & - \\ 
 \hline
 \multirow{4}{*}{32} & \textsc{mlr} &31.4&43.7&60.3&73.5\\
 & \textsc{mlp}  &32.1&44.3&61.5&74.9\\
 & \textsc{lstm}  &32.2&44.3&61.5&74.7\\
 & \textsc{cnn} &29.2&41.1&58.1&71.6\\
 \hline
 \multirow{4}{*}{64} & \textsc{mlr}  &32.1&44.9&61.9&74.3\\
 & \textsc{mlp}  &32.7&46.0&63.5&76.6\\
  & \textsc{lstm} &33.9&46.1&63.1&75.7\\

  & \textsc{cnn} &29.9&41.8&59.0&72.2\\
 \hline
  \multirow{4}{*}{96} & \textsc{mlr}  &32.0&44.5&60.7&74.6\\
 & \textsc{mlp}  &32.6&45.6&63.4&76.6\\
  & \textsc{lstm} &34.5&46.9&63.7&76.1\\
  & \textsc{cnn} &29.3&41.9&59.5&72.8\\
 \hline
  \multirow{4}{*}{128} & \textsc{mlr}  &31.7&44.5&61.0&74.3\\
 & \textsc{mlp}  &32.9&45.8&63.2&\bf 76.9\\
 & \textsc{lstm} &\bf 35.1&\bf 47.4&\bf 64.4&\bf 76.9\\
 & \textsc{cnn}  &30.2&42.3&59.6&72.8\\

\hline

\end{tabular}
\end{center}
\caption{\label{table-recall-at-k} Averaged recall at $k$ over 5 runs.}
\end{table}

\begin{table}[hbtp]
\small
\begin{center}
\tabcolsep=0.10cm
\begin{tabular}{|c|crrrrrr|}
\hline
 \multirow{2}{*}{\bf Snippet} & \multirow{2}{*}{\bf Model} & \multicolumn{3}{c}{ \bf{Frequent}} & \multicolumn{3}{c|}{\bf{Infrequent}} \\

 &&\bf F$_{we}$ &\bf R@1 & \bf R@5 &\bf F$_{we}$ & \bf R@1 & \bf R@5  \\
 \hline
 &{Base} &3.7&14.5&-&0.0&0.0&-\\ 
 \hline
 \multirow{4}{*}{32} & \textsc{mlr} &35.8&37.1&70.5&14.8&9.5&23.0\\
 & \textsc{mlp}  &35.9&38.1&71.9&13.2&9.4&21.8\\
 & \textsc{lstm}  &37.1&38.4&71.6&11.7&8.6&23.0\\
 & \textsc{cnn} &33.1&35.5&69.3&7.1&5.2&15.2\\
 \hline
 \multirow{4}{*}{64} & \textsc{mlr}  &36.7&37.9&71.8&14.9&9.9&24.0\\
 & \textsc{mlp}  &36.4&39.2&\bf 74.5&11.0&7.9&21.6\\
  & \textsc{lstm} &39.2&40.3&73.0&12.4&9.4&25.4\\

  & \textsc{cnn} &33.9&36.4&70.6&6.9&5.2&15.1\\
 \hline
  \multirow{4}{*}{96} & \textsc{mlr} &36.4&37.4&70.1&\bf 17.1&\bf 11.7&25.1\\
 & \textsc{mlp}  &36.2&39.1&74.0&11.0&7.9&23.1\\
  & \textsc{lstm} &39.6&41.1&73.7&12.4&9.6&25.8\\
  & \textsc{cnn} &32.7&35.8&71.6&6.3&4.8&13.7\\
 \hline
  \multirow{4}{*}{128} & \textsc{mlr}  &35.4&37.2&70.5&15.4&10.7&25.0\\
 & \textsc{mlp}  &36.5&39.5&74.0&11.1&8.2&22.3\\
 & \textsc{lstm} &\bf 40.3&\bf41.9&74.4&12.3&9.5&\bf 26.2\\
 & \textsc{cnn}  &33.7&36.9&71.4&6.5&5.0&14.6\\

\hline

\end{tabular}
\end{center}
\caption{\label{table-freq-infreq} Performance on \emph{frequent} (those that occur above the average) and \emph{infrequent} actions.}
\end{table}

\paragraph{Error analysis\footnote{Made over one of the runs from the \textsc{lstm}-based approach and setting the snippet size to 128 tokens. 
}} Some of the misclassifications made by the \textsc{lstm} approach were semantically related actions and counter-actions. For example, \qt{Colloportus} (to close a door) was never predicted. The most common mis-classification (14 out of 41) was \qt{Alohomora} (to unlock a door), which was 5 times more frequent in the training corpus. Similarly, \qt{Nox} (to extinguish the light from a wand) was correctly predicted 6 times, meanwhile 36 mis-classifications correspond to \qt{Lumos} (to light a place using a wand), which was 6 times more frequent in the training set. Other less frequent spells that denote vision and guidance actions, such as \qt{Point me} (the wand acts a a compass pointing North) and \qt{Homenum revelio} (to revel a human presence) were also mainly misclassified as \qt{Lumos}. This is an indicator that the \textsc{lstm} approach has difficulties to disambiguate among semantically related actions, especially if their occurrence was unbalanced in the training set. This issue is in line with the tendency observed for recall at $k$. Spells intended for much more specific purposes, according to the books, obtained a performance significantly higher than the average, e.g. F-score(\qt{Riddikulus})=63.54, F-score(\qt{Expecto Patronum})=55.49 and F-score(\qt{Obliviate})=47.45. As said before, the model is significantly biased towards frequent actions. For 79 out of 84 gold actions in the test set, we found that the samples tagged with such actions were mainly classified into one of the top 20 most frequent actions.

\paragraph{Human comparison} We collected human annotations from 208 scenes involving frequent actions. The accuracy/F-macro/F-weighted was 39.20/30.00/40.90. The \textsc{lstm} approach obtained 41.26/25.37/39.86. Overall, the \textsc{lstm} approach obtained a similar performance, but the lower macro F-score by the \textsc{lstm} could be an indicator that humans can distinguish within a wider spectrum of actions. As a side note, super-human performance it is not strange in other \textsc{nlp} tasks, such as sentiment analysis \cite{pang2002thumbs}.

\section{Conclusion}
We explored action prediction from written stories. We first introduced a corpus set in the world of Harry Potter's literature. Spells in these novels act as keywords that abstract actions. This idea was used to label a collection of fan fiction. We then evaluated standard \textsc{nlp} approaches, from logistic regression to sequential models such as \textsc{lstm}s.  The latter performed better in general, although vanilla models achieved a higher performance for actions that occurred a few times in the training set. An analysis over the output of the \textsc{lstm} approach also revealed difficulties to discriminate among semantically related actions.

The challenge here proposed corresponded to a fictional domain. A future line of work we are interested in is to test whether the knowledge learned with this dataset could be transferred to real-word actions (i.e. real-domain setups), or if such transfer is not possible and a model needs to be trained from scratch.


\section*{Acknowlegments}

This work has received support from the TELEPARES-UDC project
(FFI2014-51978-C2-2-R) and the ANSWER-ASAP project (TIN2017-85160-C2-1-R) from MINECO, and from Xunta de Galicia (ED431B 2017/01), and from the European Research Council (ERC), under the European Union's Horizon 2020 research and innovation programme (FASTPARSE, grant agreement No 714150).

\bibliographystyle{acl_natbib}
\bibliography{naaclhlt2018}

\appendix

\section{Corpus distribution}\label{appendix-corpus-details}

Table \ref{table-distribution} summarizes the label distribution across the training, development and test sets of the \textsc{hpac} corpus.

\begin{table}[!ht]
\begin{minipage}{\textwidth}
\small
\centering
\begin{tabular}{|l|r|r|r|l|r|r|r|}
\hline
\bf Action  & \bf \#Training & \bf \#Dev & \bf \#Test & \bf Action  & \bf \#Training &  \bf \#Dev & \bf \#Test \\
\hline
\textsc{avada kedavra}&7937&986&954 & \textsc{crucio}&7852&931&980 \\
\textsc{accio}&4556&595&562 & \textsc{lumos}&4159&505&531 \\
\textsc{stupefy}&3636&471&457 & \textsc{obliviate}&3200&388&397 \\
\textsc{expelliarmus}&2998&377&376 & \textsc{legilimens}&1938&237&247 \\
\textsc{expecto patronum}&1796&212&242 & \textsc{protego}&1640&196&229 \\
\textsc{sectumsempra}&1596&200&189 & \textsc{alohomora}&1365&172&174 \\
\textsc{incendio}&1346&163&186 & \textsc{scourgify}&1317&152&166 \\
\textsc{reducto}&1313&171&163 & \textsc{imperio}&1278&159&144 \\
\textsc{wingardium leviosa}&1265&158&154 & \textsc{petrificus totalus}&1253&175&134 \\
\textsc{silencio}&1145&153&136 & \textsc{reparo}&1124&159&137 \\
\textsc{muffliato}&1005&108&92 & \textsc{aguamenti}&796&84&86 \\
\textsc{finite incantatem}&693&90&75 & \textsc{incarcerous}&686&99&87 \\
\textsc{nox}&673&82&80 & \textsc{riddikulus}&655&81&88 \\
\textsc{diffindo}&565&90&82 & \textsc{impedimenta}&552&88&79 \\
\textsc{levicorpus}&535&63&68 & \textsc{evanesco}&484&53&59 \\
\textsc{sonorus}&454&66&73 & \textsc{point me}&422&57&69 \\
\textsc{episkey}&410&55&59 & \textsc{confringo}&359&52&48 \\
\textsc{engorgio}&342&52&41 & \textsc{colloportus}&269&26&41 \\
\textsc{rennervate}&253&24&33 & \textsc{portus}&238&22&31 \\
\textsc{tergeo}&235&23&26 & \textsc{morsmordre}&219&29&38 \\
\textsc{expulso}&196&23&20 & \textsc{homenum revelio}&188&30&24 \\
\textsc{mobilicorpus}&176&20&14 & \textsc{relashio}&174&20&27 \\
\textsc{locomotor}&172&24&19 & \textsc{avis}&166&17&29 \\
\textsc{rictusempra}&159&16&26 & \textsc{impervius}&149&26&13 \\
\textsc{oppugno}&144&18&7 & \textsc{furnunculus}&137&20&20 \\
\textsc{serpensortia}&133&14&15 & \textsc{confundo}&130&17&21 \\
\textsc{locomotor mortis}&127&14&15 & \textsc{tarantallegra}&126&11&17 \\
\textsc{reducio}&117&13&22 & \textsc{quietus}&108&15&17 \\
\textsc{langlock}&99&12&19 & \textsc{geminio}&78&5&10 \\
\textsc{ferula}&78&6&10 & \textsc{orchideous}&76&7&5 \\
\textsc{densaugeo}&67&13&8 & \textsc{liberacorpus}&63&7&5 \\
\textsc{aparecium}&63&14&10 & \textsc{anapneo}&62&6&5 \\
\textsc{flagrate}&59&4&11 & \textsc{deletrius}&59&12&6 \\
\textsc{obscuro}&57&11&7 & \textsc{prior incantato}&56&4&3 \\
\textsc{deprimo}&51&2&2 & \textsc{specialis revelio}&50&11&6 \\
\textsc{waddiwasi}&45&5&8 & \textsc{protego totalum}&44&9&5 \\
\textsc{duro}&36&4&4 & \textsc{salvio hexia}&36&8&5 \\
\textsc{defodio}&34&2&6 & \textsc{piertotum locomotor}&30&4&3 \\
\textsc{glisseo}&26&4&3 & \textsc{mobiliarbus}&25&3&4 \\
\textsc{repello muggletum}&23&2&5 & \textsc{erecto}&23&7&5 \\
\textsc{cave inimicum}&19&5&2 & \textsc{descendo}&19&0&1 \\
\textsc{protego horribilis}&18&7&5 & \textsc{meteolojinx recanto}&10&3&1 \\
\textsc{peskipiksi pesternomi}&7&0&0 &&&&\\

\hline

\end{tabular}
\caption{\label{table-distribution} Label distribution for the \textsc{hpac} corpus}
\end{minipage}
\end{table}



\end{document}